\documentclass[10pt,twocolumn,letterpaper]{article}

\usepackage{iccv}
\usepackage{times}
\usepackage{epsfig}
\usepackage{graphicx}
\usepackage{amsmath}
\usepackage{amssymb}

\usepackage{booktabs}
\usepackage[linesnumbered,ruled,vlined]{algorithm2e}
\SetKwInput{KwInput}{Input}
\SetKwInput{KwOutput}{Output}
\usepackage{commath}
\usepackage{caption}
\usepackage{subcaption}
\usepackage{verbatim}
\usepackage{gensymb}
\usepackage{rotating}
\usepackage{marvosym}
\usepackage{multirow}
\usepackage{bm}
\usepackage{enumitem}
\usepackage[dvipsnames]{xcolor}

\usepackage[pagebackref=true,breaklinks=true,letterpaper=true,colorlinks,bookmarks=false]{hyperref}

\iccvfinalcopy 

\def\iccvPaperID{0005} 

\begin{document}

\title{Generative Approach for\\Probabilistic Human Mesh Recovery using Diffusion Models}

\author{Hanbyel Cho\\
School of Electrical Engineering, KAIST\\
{\tt\small \texttt{tlrl4658@kaist.ac.kr}}
\and
Junmo Kim\\
School of Electrical Engineering, KAIST\\
{\tt\small \texttt{junmo.kim@kaist.ac.kr}}
}

\maketitle

\begin{abstract}
\vspace{-2mm}
    This work focuses on the problem of reconstructing a 3D human body mesh from a given 2D image. Despite the inherent ambiguity of the task of human mesh recovery, most existing works have adopted a method of regressing a single output. In contrast, we propose a generative approach framework, called ``Diffusion-based Human Mesh Recovery (Diff-HMR)'' that takes advantage of the denoising diffusion process to account for multiple plausible outcomes. During the training phase, the SMPL parameters are diffused from ground-truth parameters to random distribution, and Diff-HMR learns the reverse process of this diffusion. In the inference phase, the model progressively refines the given random SMPL parameters into the corresponding parameters that align with the input image. Diff-HMR, being a generative approach, is capable of generating diverse results for the same input image as the input noise varies. We conduct validation experiments, and the results demonstrate that the proposed framework effectively models the inherent ambiguity of the task of human mesh recovery in a probabilistic manner. Code is available at \url{https://github.com/hanbyel0105/Diff-HMR}.
\end{abstract}
\vspace{-2mm}
\section{Introduction}
\vspace{-1mm}
Human Mesh Recovery (HMR) is a task of regressing three-dimensional human body model parameters, such as SMPL~\cite{ref1_SMPL}, from a given 2D image. Along with joint-based methods~\cite{
ref_pavllo20193d,ref_CamDistHumanPose3D,ref_liu2020attention}, HMR is a fundamental task in computer vision and holds great significance in applications like computer graphics and VR/AR. Despite significant advances in HMR, it remains a challenging problem due to the inherent ambiguity caused by the loss of depth information and occlusions in 2D images. Most existing approaches~\cite{ref3_HMR,ref4_SPIN,ref_HMR-ViT,ref_pymaf,ref_imphmr,ref_hmr_survey} have relied on single-output regression, limiting their ability to explain uncertainty in the HMR process. Consequently, these methods often fail to reconstruct diverse and plausible 3D human body meshes that accurately represent the actual posture of the human.

    \begin{figure}[!t]
    \centering
    \vspace{-2mm}
        \includegraphics[width=\columnwidth]{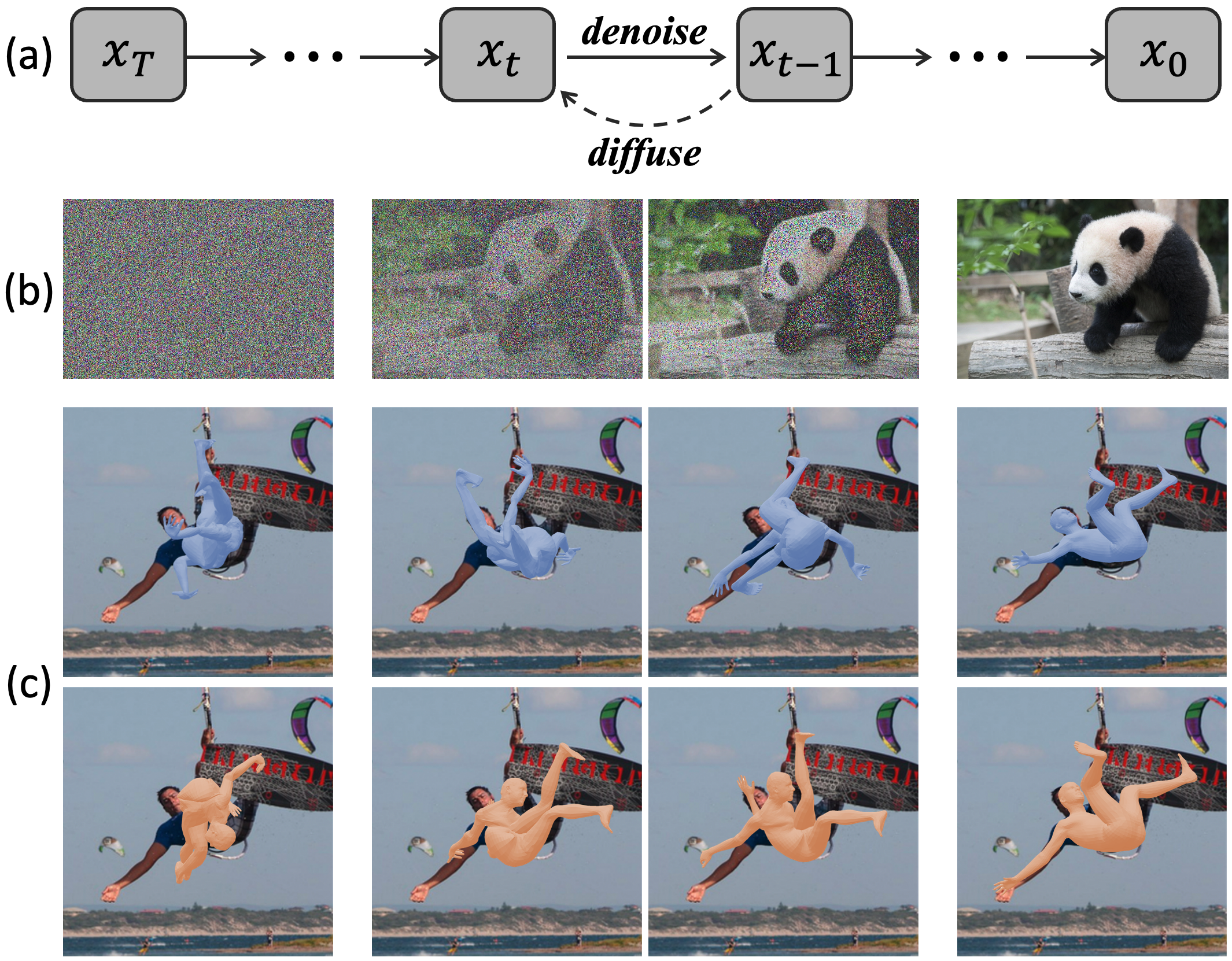}
        \caption{\textbf{Diffusion model for human mesh recovery.} (a) A diffusion model defined by a Markov chain, consisting of a forward process (``\emph{diffuse}'') and a reverse process (``\emph{denoise}''). (b) Diffusion model for the task of image generation. (c) We propose Diff-HMR, which formulates human mesh recovery as a denoising diffusion process, enabling the generation of multiple plausible meshes based on variations in a random noise input of SMPL.}
        \label{fig:teaser}
        \vspace{-1mm}
    \end{figure}

To overcome these limitations, we propose a generative approach framework called ``Diffusion-based Human Mesh Recovery (\emph{Diff-HMR})''. Unlike traditional \emph{perception}-based approaches~\cite{ref3_HMR,ref4_SPIN,ref_pymaf,ref_hybrik} that regress a single output from the given image, our proposed framework draws inspiration from the denoising diffusion process~\cite{ref_DDPM} and outputs SMPL parameters in a \emph{generative} way from random noise input. By exploring multiple plausible outcomes during the reconstruction process based on the input noise varying, Diff-HMR can probabilistically model the inherent ambiguity in human mesh recovery.

Specifically, during the training phase, the SMPL parameters are diffused from ground-truth parameters to random distribution, and Diff-HMR learns the reverse of this diffusion process. In the inference phase, the model progressively refines the initial random SMPL parameters to the parameters corresponding to the given input image. As a generative approach, Diff-HMR can produce diverse plausible human body meshes for the same input image as the input noise varies as shown in Figure~\ref{fig:teaser} (c).

Additionally, to overcome the limitation of the diffusion model, which is sensitive to signal-to-noise ratio (SNR), we adopt 6D rotation representations~\cite{ref_6d_rot} instead of Axis-angle representations as joint angle representations for SMPL. We conduct experiments on various datasets~\cite{ref_3dpw,ref_COCO} and verify the efficacy of Diff-HMR. Our contributions are as follows:
    \begin{itemize}[leftmargin=*]
        \vspace{-0.8mm}
        \item
        We propose a novel generative HMR framework, Diff-HMR, which utilizes the denoising diffusion process to generate multiple 3D human meshes from a given image, effectively modeling the inherent ambiguity of HMR.
        \vspace{-0.8mm}
        
        \item
        By adopting 6D rotation representations for SMPL joint angles, Diff-HMR can maintain a suitable signal-to-noise ratio (SNR), enabling stable training of diffusion models.
        \vspace{-4.3mm}
        
        \item
        The experimental results show that Diff-HMR effectively models the inherent ambiguity of HMR, generating multiple plausible outcomes from a given image.
    \end{itemize}

    \begin{figure*}[!t]
    \centering
        \includegraphics[width=0.99\linewidth]{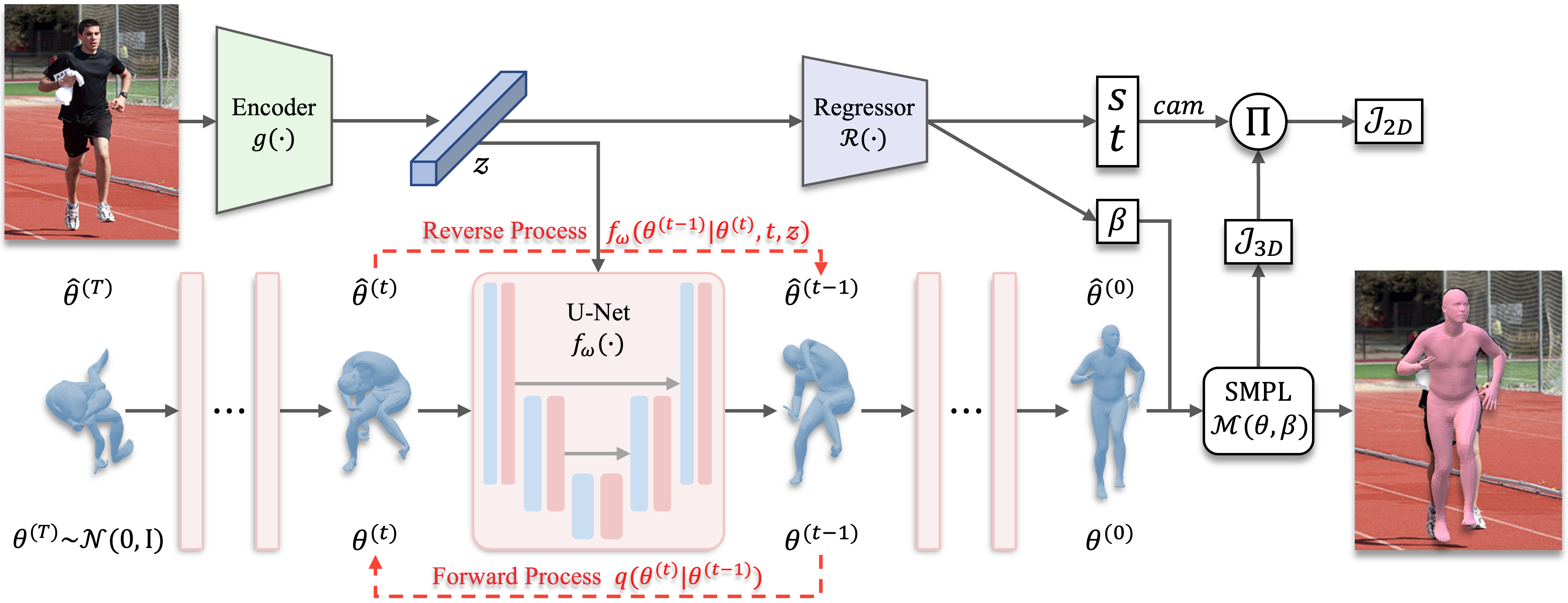}
        \vspace{-0.5mm}
        \caption{\textbf{Illustration of Diff-HMR architecture.}
        For a given image, Diff-HMR can predict multiple plausible human body meshes that a person in the image may take. To model the ambiguity of the posture taken by the person, Diff-HMR progressively generates the final human mesh $\boldsymbol{\hat{\theta}}^{(0)}$ from the noise SMPL pose parameters $\boldsymbol{\hat{\theta}}^{(T)}$ that follows $\mathcal{N}(\mathbf{0}, \mathbf{I})$. During the \emph{reverse process}, the denoising network $f_{\boldsymbol{\omega}}(\cdot)$ denoises $\boldsymbol{\hat{\theta}}^{(T)}$ as the correct SMPL pose $\boldsymbol{\hat{\theta}}^{(0)}$, referring to the feature vector $\boldsymbol{z}$ extracted from the input image as $f_{\boldsymbol{\omega}}(\boldsymbol{\theta}^{(t-1)}|\boldsymbol{\theta}^{(t)},t,\boldsymbol{z})$. In order to train $f_{\boldsymbol{\omega}}(\cdot)$, the \emph{forward process} gradually perturbs the ground-truth SMPL pose $\boldsymbol{\theta}^{(0)}$ as $q(\boldsymbol{\theta}^{(t)}|\boldsymbol{\theta}^{(t-1)})$ to produce the noised SMPL pose $\boldsymbol{\theta}^{(t)}$ required for training.
        }
        \vspace{-1.5mm}
        \label{fig:overall}
    \end{figure*}
\section{Related Work}
    \paragraph{Monocular 3D Human Mesh Recovery} can be categorized into optimization-based, regression-based, and hybrid approaches. Optimization-based approaches~\cite{ref2_SMPLify,rw_ref2_unite} fit SMPL parameters to minimize errors between reconstructed meshes and 3D/2D evidence like keypoints or silhouettes. Regression-based approaches~\cite{ref3_HMR,ref15_NeuralBodyFit,ref_HMR-ViT,ref13_LearningToEstim,ref_imphmr,ref_pymaf} leverage deep neural networks to improve inference speed and reconstruction quality by directly inferring SMPL from input images. In recent studies, hybrid approaches~\cite{ref4_SPIN,ref_eft} that combine both optimization and regression-based methods have emerged. This approach offers a more accurate pseudo-ground-truth SMPL for 2D images. Despite these advancements, such single-output regression methods still limit their ability to explain the uncertainty in the HMR. To overcome this, we propose Diff-HMR that produces multiple plausible outcomes from random noise of SMPL, which provides comprehensive modeling in probabilistic manner.

    \vspace{-3mm}
    \paragraph{Probabilistic Inference for Human Body Mesh.}
        Estimating a person’s 3D pose from a single image remains challenging due to inherent ambiguities such as occlusions and depth ambiguities. To address this problem, Jahangiri~\etal\cite{ref_rw_Jahangiri} adopted a compositional model and utilized anatomical evidence to infer multiple hypotheses of 3D poses. In addition, Li~\etal\cite{ref_rw_Li} used a Mixture Density Network, considering the Gaussian kernel centroids as individual hypotheses. Sharma~\etal\cite{ref_rw_Sharma} utilized CVAE~\etal\cite{ref_rw_CVAE} to generate hypotheses and generated the final output by using the weighted average of hypotheses based on joint-ordinal relations. In the field of human mesh recovery, following this trend, recent works such as Biggs~\etal\cite{ref_rw_Biggs} have utilized Normalizing flow to output N pre-defined predictions, and Kolotouros~\etal\cite{ref6_ProHMR} directly outputs likelihood values. However, despite the excellent capabilities of a denoising diffusion process for probability distribution modeling, it has not yet been applied to probabilistic HMR.
        
    \vspace{-4mm}
    \paragraph{Denoising Diffusion Probabilistic Models.}
        Denoising diffusion probabilistic models (DDPMs)~\cite{ref_DDPM}, also known as diffusion models, are generative models composed of two stages: a forward process (``\emph{diffuse}'') and a reverse process (``\emph{denoise}''). In the forward process, the input data is diffused into a random distribution through multiple steps by adding Gaussian noise. In the reverse process, the model learns to reverse the forward process, denoising the noised data, and consequently learns the distribution of data. DDPMs have shown excellent performance in density estimation and have recently been applied to tasks such as image/text generation~\cite{ref_rw_ddpm_gen_1,ref_rw_ddpm_gen_2,ref_rw_ddpm_gen_3,ref_rw_ddpm_gen_4,ref_rw_ddpm_gen_5,ref_rw_ddpm_gen_6_text}, despite the computational cost of the sampling process. In the context of joint-based 3D human pose estimation, DDPMs have been applied to tackle ambiguities~\cite{ref_rw_diffpose_cvpr23,ref_rw_D3DP_iccv23}. However, there is no method for human mesh recovery yet; therefore, in this work, we propose a probabilistic HMR method that considers multiple outputs by applying DDPMs for the first time.
\section{Method}
\vspace{-1mm}
The overall framework of our Diff-HMR is depicted in Figure~\ref{fig:overall}. In this section, we first recapitulate DDPMs~\cite{ref_DDPM} and then explain the model architecture of Diff-HMR.
\vspace{-1mm}
\subsection{Preliminary: DDPMs}
\vspace{-1mm}
Denoising diffusion probabilistic models (DDPMs) are composed of two stage, each defined as a Markov chain: the \emph{forward process} and the \emph{reverse process}.
\vspace{-4mm}
\paragraph{Forward Process.} Let $\boldsymbol{x}^{(0)}$ be the observed original data, and it follows an unknown distribution $q(\boldsymbol{x}^{(0)})$. The goal of the forward process, referred to as the diffusion process, is to transform $\boldsymbol{x}^{(0)}$ into a Gaussian distribution $\mathcal{N}(\mathbf{0}, \mathbf{I})$ by gradually adding Gaussian noise over pre-defined $T$ steps.

At each step $t$, the data is progressively disturbed by adding noise according to the following equation:
\vspace{-0.3mm}
    \begin{align}
        \begin{split}\label{eq:1}
            q(\boldsymbol{x}^{(t)} | \boldsymbol{x}^{(t-1)}) \sim \mathcal{N}(\sqrt{1-\beta_t}\boldsymbol{x}^{(t-1)}, \beta_t\mathbf{I})
        \end{split}
    \end{align}
where $t=1,\cdots,T$ and $\beta_t$ is pre-defined noise schedule; We can rewrite $\boldsymbol{x}^{(t)}$ as $\boldsymbol{x}^{(t)} = \sqrt{1-\beta_t}\boldsymbol{x}^{(t-1)}+\sqrt{\beta_t}\boldsymbol{\epsilon}_t$, where $\boldsymbol{\epsilon}_t$ follows a Gaussian distribution $\mathcal{N}(\mathbf{0}, \mathbf{I})$.

For any given $t$, since $\boldsymbol{\epsilon}_t$ are \emph{i.i.d.}, we can directly generate $\boldsymbol{x}^{(t)}$ from $\boldsymbol{x}^{(0)}$ in a closed form as follows: 
    \begin{align}
        \begin{split}\label{eq:2}
            q(\boldsymbol{x}^{(t)} | \boldsymbol{x}^{(0)}) \sim \mathcal{N}(\sqrt{\Bar{\alpha}_t}\boldsymbol{x}_0, (1-\Bar{\alpha}_t)\mathbf{I})
        \end{split}
    \end{align}
where $\alpha_t := 1-\beta_t$ and $\Bar{\alpha}_t := \prod_{s=1}^t \alpha_s$.

By using Eq.~\ref{eq:2}, we can efficiently generate noised samples $\boldsymbol{x}^{(t)}$ during the training phase. In Diff-HMR, we employ ground-truth SMPL pose parameters $\boldsymbol{\theta}^{(0)}$ as $\boldsymbol{x}^{(0)}$ to get the noised SMPL pose $\boldsymbol{\theta}^{(t)}$ in the forward process.
\vspace{-4mm}\paragraph{Reverse Process.} Let us define the joint distribution $p_{\boldsymbol{\omega}}(\boldsymbol{x}^{(0:T)})$, parameterized by the learnable parameters $\boldsymbol{\omega}$, which aims to mimic the distribution $q(\boldsymbol{x}^{(0)})$ as follows:
    \begin{align}
        \begin{split}\label{eq:3}
            p_{\boldsymbol{\omega}}(\boldsymbol{x}^{(0:T)}) = p(\boldsymbol{x}^{(T)})\prod_{t=1}^{T}p_{\boldsymbol{\omega}}(\boldsymbol{x}^{(t-1)} | \boldsymbol{x}^{(t)})
        \end{split}
    \end{align}
where 
$p_{\boldsymbol{\omega}}(\boldsymbol{x}^{(t-1)} | \boldsymbol{x}^{(t)}) \sim \mathcal{N}(\boldsymbol{\mu}_{\boldsymbol{\omega}}(\boldsymbol{x}^{(t)}, t), \boldsymbol{\Sigma}_{\boldsymbol{\omega}}(\boldsymbol{x}^{(t)}, t))$ and $p(\boldsymbol{x}^{(T)}) \sim \mathcal{N}(\mathbf{0}, \mathbf{I})$. The goal of the reverse process is to find $\boldsymbol{\omega}$ that maximizes $p(\boldsymbol{x}^{(0)})$ when $\boldsymbol{x}^{(0)}$ follows $q(\boldsymbol{x}^{(0)})$.

To achieve this, we need to learn $\boldsymbol{\mu}_{\boldsymbol{\omega}}(\boldsymbol{x}^{(t)}, t)$ (\emph{mean}) and $\boldsymbol{\Sigma}_{\boldsymbol{\omega}}(\boldsymbol{x}^{(t)}, t)$ (\emph{variance}). However, according to DDPMs, variance can be expressed as $\boldsymbol{\Sigma}_{\boldsymbol{\omega}}(\boldsymbol{x}^{(t)}, t) = \beta_t \frac{1-\Bar{\alpha}_{t-1}}{1-\Bar{\alpha}_{t}} \mathbf{I}$, which only depends on $t$; Thus we only need to learn mean.

Furthermore, since the mean can be reparameterized as $\boldsymbol{\mu}_{\boldsymbol{\omega}}(\boldsymbol{x}^{(t)}, t) = \frac{1}{\sqrt{\alpha_t}} ( \boldsymbol{x}^{(t)}-\frac{\beta_t}{\sqrt{1 - \Bar{\alpha}_t}}\boldsymbol{\epsilon}_{\boldsymbol{\omega}}(\boldsymbol{x}^{(t)}, t) )$, we can simply train a neural network $\boldsymbol{\epsilon}_{\boldsymbol{\omega}}(\boldsymbol{x}^{(t)}, t)$ parameterized by $\boldsymbol{\omega}$ to predict the noise $\boldsymbol{\epsilon}$ at each reverse step.

In Diff-HMR, the model extracts a feature vector $\boldsymbol{z}$ from the input image and predicts conditioned noise $\boldsymbol{\epsilon}$ on the $\boldsymbol{z}$ to generate the pose corresponding to the person of the input image. Therefore, the training objective for the reverse process in Diff-HMR is $\mathbb{E} [ \lVert \boldsymbol{\epsilon}-\boldsymbol{\epsilon}_{\boldsymbol{\omega}}(\boldsymbol{x}^{(t)}, t, \boldsymbol{z}) \rVert^2 ]$, where $\boldsymbol{\epsilon} \sim \mathcal{N}(\mathbf{0}, \mathbf{I})$ and $\boldsymbol{x}^{(t)}$ is as the following equation:
    \begin{align}
        \begin{split}\label{eq:4}
            \boldsymbol{x}^{(t)} = \sqrt{\Bar{\alpha}_{t}}\boldsymbol{x}^{(0)}+\sqrt{1-\Bar{\alpha}_{t}}\boldsymbol{\epsilon}.
        \end{split}
    \end{align}

\subsection{Diffusion-based Human Mesh Recovery}
\vspace{-0.5mm}
In this work, we propose a diffusion-based human mesh recovery (\emph{Diff-HMR}) by incorporating denoising diffusion probabilistic models (DDPMs)~\cite{ref_DDPM} into the existing HMR framework to address the inherent ambiguity of the HMR.

\vspace{-4.5mm}\paragraph{Overall Framework.} To tackle the inherent ambiguity of human poses in the given image, Diff-HMR gradually generates the correct human mesh $\boldsymbol{\hat{\theta}}^{(0)}$ corresponding to the image from noise SMPL pose parameters $\boldsymbol{\hat{\theta}}^{(T)}$ as in the reverse process in DDPMs. To this end, we first extract the image feature $\boldsymbol{z}$ from the given image $\textbf{I}$ using the ResNet-50~\cite{ref_resnet}-based image encoder $g(\cdot)$ as $\boldsymbol{z} = g(\textbf{I}) \in \mathbb{R}^{2048 \times 7 \times 7}$.

At each reverse step $t$, the denoising network $f_{\boldsymbol{\omega}}(\cdot)$ with 1D U-Net~\cite{ref_unet} architecture is conditioned on the time step $t$ and image feature $\boldsymbol{z}$ to predict the noise $\boldsymbol{\hat{\epsilon}}_t$ necessary to estimate the previous pose $\boldsymbol{\hat{\theta}}^{(t-1)}$ from $\boldsymbol{\hat{\theta}}^{(t)}$. To train $f_{\boldsymbol{\omega}}(\cdot)$, the ground-truth noise $\boldsymbol{\epsilon}_t$ is generated from the ground-truth SMPL pose $\boldsymbol{\theta}^{(0)}$ through the forward process using Eq.~\ref{eq:4}. We define loss function for typical DDPMs training that predicts the noise $\boldsymbol{\hat{\epsilon}}_t$ as $\mathcal{L}_\text{diff}=\mathbb{E} [ \lVert \boldsymbol{\epsilon}_t-f_{\boldsymbol{\omega}}(\boldsymbol{\theta}^{(t)}, t, \boldsymbol{z}) \rVert^2 ]$ and use this for training of the denoising network $f_{\boldsymbol{\omega}}(\cdot)$.

\vspace{-4mm}\paragraph{Training Objective.}
In addition to using $\mathcal{L}_\text{diff}$, we also impose constraints on the final SMPL pose $\boldsymbol{\hat{\theta}}^{(0)}$ and the derived 2D ({\small $\mathcal{J}_\text{2D}$}) and 3D ({\small $\mathcal{J}_\text{3D}$}) joints as in conventional HMR methods~\cite{ref3_HMR,ref_imphmr}, which are represented as $\mathcal{L}_\text{hmr}$. However, since $f_{\boldsymbol{\omega}}(\cdot)$ predicts the noise at each time step $t$, we cannot directly access $\boldsymbol{\hat{\theta}}^{(0)}$. To overcome this, we deduce $\boldsymbol{\hat{\theta}}^{(0)}$ from the predicted noise $\boldsymbol{\hat{\epsilon}}_t$ by reparameterizing Eq.~\ref{eq:4} as $\boldsymbol{\hat{\theta}}^{(0)} = \frac{1}{\sqrt{\bar{\alpha_t}}}\boldsymbol{\hat{\theta}}^{(t)} - (\sqrt{\frac{1}{\bar{\alpha_t}} - 1})f_{\boldsymbol{\omega}}(\boldsymbol{\hat{\theta}}^{(t)}, t, \boldsymbol{z})$. Thus, our overall loss function is $\mathcal{L}_\text{all}=\mathcal{L}_\text{diff}+\mathcal{L}_\text{hmr}$.

Since the ambiguity of human mesh recovery is mainly related to pose, Diff-HMR predicts SMPL shape parameters $\boldsymbol{\beta}$ and weak-perspective camera parameters $\bm{\pi} \in [s (\emph{scale}),t (\emph{translation})]$ simply using MLP-based regressor $\mathcal{R}(\cdot)$ as $\{\boldsymbol{\beta}, \bm{\pi}\} = \mathcal{R}(\boldsymbol{z})$. From predicted results, we can get mesh vertices $M=\mathcal{M}(\bm{\theta}, \bm{\beta}) \in \mathbb{R}^{6890 \times 3}$ where $\mathcal{M}$ is transform function and get 3D joints $\mathcal{J}_\text{3D} = WM$ and 2D joints $\mathcal{J}_\text{2D} = \bm{\Pi}(\mathcal{J}_\text{3D})$ where $W$ and $\bm{\Pi}$ denote pre-trained linear regressor and projection function, respectively.
    \begin{figure}[!t]
    \centering
        \includegraphics[width=\columnwidth]{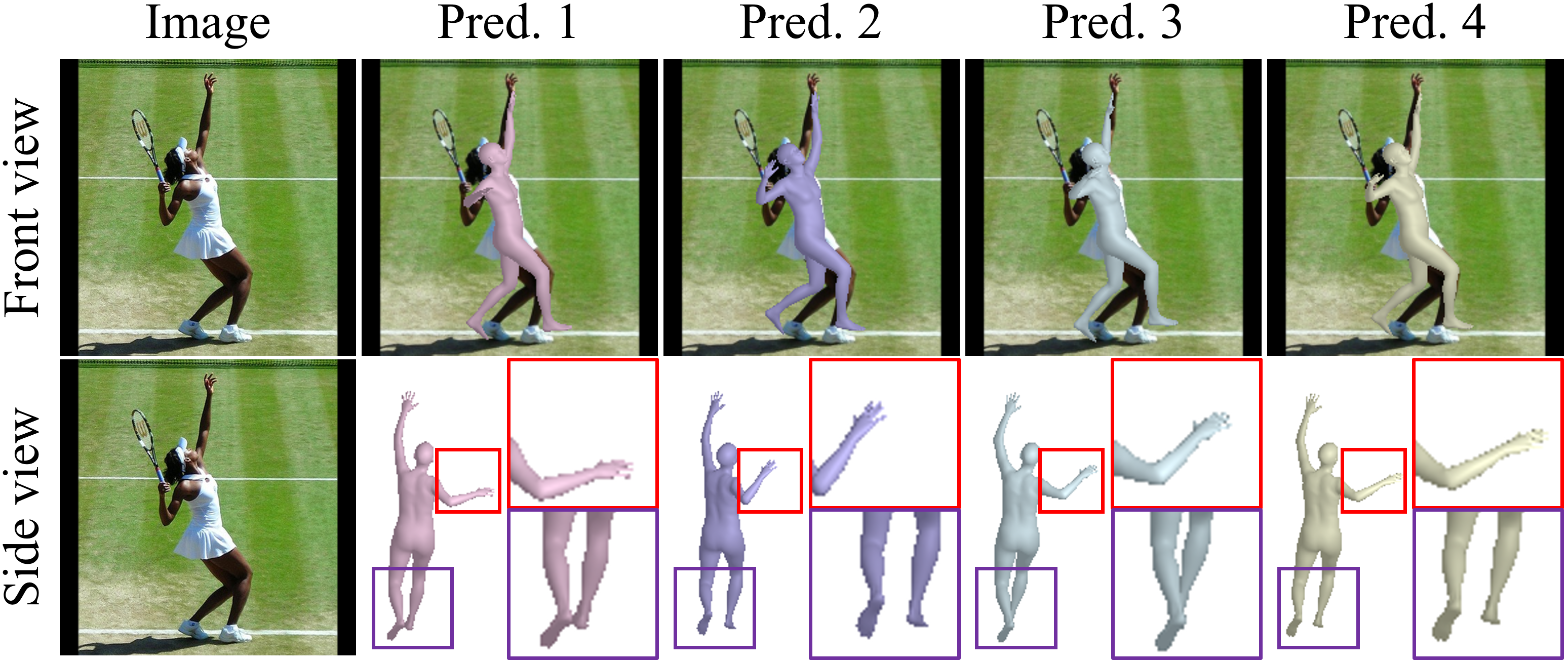}
        \vspace{-6mm}
        \caption{\textbf{Predicted meshes obtained by varying seeds.}
        Diff-HMR outputs a plausible human body mesh of different possibilities for each sampled seed {\small $\boldsymbol{\hat{\theta}}^{(T)} \sim \mathcal{N}(\mathbf{0}, \mathbf{I})$}.
        }
        \label{fig:qualitative}
    \end{figure}

    \begin{table}[!t]
    \centering
    \resizebox{\columnwidth}{!}{
    \begin{tabular}{@{}l|c|c|c@{}}
        \toprule
		    ~~~\bf Method & {MPJPE $\downarrow$} & {PA-MPJPE $\downarrow$} & {PVE $\downarrow$} ~~\\
		\midrule
            ~~ SPIN~\cite{ref4_SPIN} ICCV'19                & 96.9 & 59.2 & 116.4 ~~\\
            ~~ Biggs~\etal\cite{ref_rw_Biggs} NeurIPS'20    & {\small N/A} & 59.9 & {\small N/A} ~~\\
            ~~ ProHMR~\cite{ref6_ProHMR} ICCV'21            & {\small N/A} & 59.8 & {\small N/A} ~~\\
		\midrule
            ~~ Diff-HMR, $n=1$  & 98.9 & 58.5 & 114.6 ~~\\
            ~~ Diff-HMR, $n=5$  & 96.3 & 57.0 & 111.8 ~~\\
            ~~ Diff-HMR, $n=10$ & \underline{95.5} & \underline{56.5} & \underline{110.9} ~~\\
            ~~ Diff-HMR, $n=25$ & \textbf{94.5} & \textbf{55.9} & \textbf{109.8} ~~\\
		\bottomrule
	\end{tabular}
    }
    \vspace{-1.7mm}
    \caption{\textbf{Multiple hypotheses results on 3DPW.}
    Values are in mm. $n$ denotes the number of inferences performed by varying the seed {\small $\boldsymbol{\hat{\theta}}^{(T)}$}. We report the error calculated based on the minimum error among $n$ samples. Best in bold, second-best underlined.
    }
    \label{tab:quant}
    \vspace{-3mm}
    \end{table}{}

\section{Experiments}
\vspace{-1mm}
\subsection{Datasets and Evaluation Metrics}
\vspace{-1mm}
In order to train our model, we use H36M~\cite{ref_h36m} and MPI-INF-3DHP~\cite{ref_mpii3d} as 3D datasets and pseudo-ground-truth SMPL annotated COCO~\cite{ref_COCO} and MPII~\cite{ref_mpii2d} as 2D datasets. For quantitative evaluation, we use 3DPW~\cite{ref_3dpw} test split and use the Mean Per Joint Position Error (MPJPE), Procrustes-Aligned MPJPE (PA-MPJPE), and Per-Vertex Error (PVE) as our evaluation metrics. For qualitative evaluation, we use the sample image from the validation set of COCO~\cite{ref_COCO}.

\vspace{-1mm}
\subsection{Experimental Results}
\vspace{-1mm}
This section includes basic experiments to validate Diff-HMR. We show quantitative and qualitative results and explore the efficacy of using 6D rotation representations as joint angle representations for SMPL in diffusion models.

\vspace{-4mm}\paragraph{Quantitative results.}
Table~\ref{tab:quant} shows quantitative results of Diff-HMR. When $n$ is $1$, Diff-HMR shows comparable performance to existing HMR methods~\cite{ref4_SPIN,ref_rw_Biggs,ref6_ProHMR}. Moreover, as $n$ increases to $5$, $10$, and $25$, the errors decrease significantly, indicating that Diff-HMR has strong representational power for modeling the inherent ambiguity of HMR.

\vspace{-4mm}\paragraph{Qualitative results.}
Figure~\ref{fig:qualitative} shows human mesh reconstruction results of Diff-HMR. As the input noise SMPL pose parameters $\boldsymbol{\hat{\theta}}^{(T)}$ vary along Gaussian distribution $\mathcal{N}(\mathbf{0}, \mathbf{I})$, the model exhibits diverse inference outcomes for probabilistically ambiguous body parts.

\vspace{-4mm}\paragraph{Efficacy of using 6D rotation representations.}
For stable training of diffusion models~\cite{ref_DDPM}, the ratio between the magnitude of the original data and the injected noise, known as the signal-to-noise ratio (SNR), is crucial. To incorporate diffusion models into HMR, we adopt 6D rotation representations~\cite{ref_6d_rot} with the same scale as Gaussian distribution as joint angle representations of SMPL pose parameters. As shown in Table~\ref{tab:ablation}, we can confirm that it is more reasonable to adopt 6D rotation than Axis-angle representations.

    \begin{table}[!t]
    \centering
    \resizebox{0.98\columnwidth}{!}{
    \begin{tabular}{@{}l|c|c@{}}
        \toprule
            ~~~\bf Method                           ~&~ 3DPW~\cite{ref_3dpw}    ~&~ MPI-INF-3DHP~\cite{ref_mpii3d}  ~~~\\
		\midrule
            ~~ Axis-angle repr.                     ~&~ 116.3                   ~&~ 146.0                           ~~~\\
            ~~ 6D rotation repr.~\cite{ref_6d_rot}  ~&~ \textbf{64.5}           ~&~ \textbf{92.5}                   ~~~\\
		\bottomrule
	\end{tabular}
    }
    \vspace{-1mm}
    \caption{\textbf{Performance comparison based on SMPL joint angle representations.}
    We report PA-MPJPE in mm. Diff-HMR shows better results when adopting 6D rotation representations as joint angle representations. Note that only the COCO~\cite{ref_COCO} is used for training here. Best in bold.
    }
    \vspace{-1mm}
    \label{tab:ablation}
    \end{table}{}
\section{Conclusion and Future Work}
\vspace{-1mm}\paragraph{Conclusion.} In this work, we focus on the task of human mesh recovery, which reconstructs 3D human body mesh from a given 2D observation. To probabilistically model the inherent ambiguity of the task, we propose a generative approach framework, called ``\emph{Diff-HMR}'' that takes advantage of the denoising diffusion process to account for multiple plausible outcomes. To overcome the limitation of the diffusion model, which is sensitive to the signal-to-noise ratio (SNR), we adopt 6D rotation representations instead of Axis-angle representations as joint angle representations for SMPL. The evaluations on benchmark datasets demonstrate that the proposed framework successfully models the inherent ambiguity of the task of human mesh recovery.
\vspace{-4mm}\paragraph{Future Work.} We will focus on improving the conditioning module of the 1D U-Net to better understand the spatial context of the input image. By enhancing the module, we can expect Diff-HMR to infer more plausible results, especially in occluded situations.

{\small
\bibliographystyle{ieee_fullname}
\bibliography{egbib}
}

\end{document}